\pdfoutput=1
\documentclass[11pt]{article}

\usepackage[]{acl}

\usepackage{times}
\usepackage{latexsym}
\usepackage{multirow}
\usepackage{graphicx} 
\usepackage{enumitem}
\usepackage{placeins}
\usepackage{comment}
\usepackage{caption}
\usepackage{subcaption}
\usepackage{placeins}

\usepackage[T1]{fontenc}
\usepackage[utf8]{inputenc}

\usepackage{microtype}

\usepackage{inconsolata}

\usepackage{booktabs}

\title{What Drives Performance in Multilingual Language Models? \\ }

\author{Sina Bagheri Nezhad, Ameeta Agrawal \\
  Portland State University \\
  \texttt{\{sina.bagherinezhad, ameeta\}@pdx.edu} \\}

\begin{document}
\maketitle
\begin{abstract}
This study investigates the factors influencing the performance of multilingual large language models (MLLMs) across diverse languages. We study 6 MLLMs, including masked language models, autoregressive models, and instruction-tuned LLMs, on the SIB-200 dataset, a topic classification dataset encompassing 204 languages. Our analysis considers three scenarios: ALL languages, SEEN languages (present in the model's pretraining data), and UNSEEN languages (not present or documented in the model's pretraining data in any meaningful way). We examine the impact of factors such as pretraining data size, general resource availability, language family, and script type on model performance. Decision tree analysis reveals that pretraining data size is the most influential factor for SEEN languages. However, interestingly, script type and language family are crucial for UNSEEN languages, highlighting the importance of cross-lingual transfer learning. 
Notably, model size and architecture do not significantly alter the most important features identified. Our findings provide valuable insights into the strengths and limitations of current MLLMs and hope to guide the development of more effective and equitable multilingual NLP systems.\footnote{\url{https://github.com/PortNLP/MLLMs_performance}}
\end{abstract}

%\sina{one of our important observations was that the instruction-tuned models (BLOOMZ) is more sensitive to pretraining data distribution (BLOOM dataset) than its fine-tuning dataset.}\mee{this makes perfect sense because pretraining data is much larger than fine-tuning, and because they're both similar distributions anyway. I don't think we need to highlight this in the abstract}\sina{Ok, Thanks.} % Interestingly, instruction-tuned models like BLOOMZ remain heavily influenced by pretraining data distribution. 

\section{Introduction}

Multilingual large language models (MLLMs) have revolutionized natural language processing by enabling applications like machine translation and sentiment analysis across numerous languages \cite{barbieri-etal-2022-xlm, yang-etal-2023-BigTrans}. Understanding how these models perform across languages with diverse linguistic properties is crucial for further development \cite{devlin-etal-2019-bert, wu2020all, scao2022bloom, lai2023chatgpt, ahuja2023mega}. Despite significant progress, linguistic disparities persist in NLP, highlighting the need for models that perform effectively and safely across a wider range of languages \cite{joshi-etal-2020-state,ranathunga2022some, 10302481,wang2023all}.

The factors contributing to the effectiveness of MLLMs, however, remain unclear. While several studies suggest the amount of language-specific pretraining data as a key factor \cite{wu2020all, scao2022bloom, shliazhko2022mgpt, ahuja2023mega}, most investigations are limited in scope, focusing on a small set of languages, specific tasks, or training paradigms like masked language modeling (MLM) or autoregressive models. {Crucially, prior work often overlooks the distinction between languages encountered during pretraining (SEEN), languages entirely new to the model (UNSEEN), and the complete set of languages available in the evaluation dataset (ALL). The question remains -- {\em what factors are important in the case of unseen languages where language-specific pretraining data is not one of the relevant factors?} This distinction is essential for understanding how MLLMs generalize to languages with varying levels of familiarity.}

\begin{table*}[ht]
\centering
\resizebox{\textwidth}{!}{
\begin{tabular}{l|p{6.5cm}|p{4cm}|c}
\toprule
\textbf{Reference} & \textbf{Factors} & \textbf{Task} & \textbf{Languages} \\
\midrule
\citet{wu2020all} & Pretraining data size, Task-specific data size, Vocabulary size &  NER & 99 \\

\citet{scao2022bloom} & Pretraining data size, Task-specific data size, Language family, Language script  & Probing & 17\\

\citet{shliazhko2022mgpt} & Pretraining data size, Language script, Model size & Perplexity & 61 \\

\citet{ahuja2023mega} & Pretraining data size, Tokenizer fertility  & Classification, QA, Sequence Labeling, NLG & 2-48 \\

Ours & Pretraining data size, Language family, {Language script,  General resource availability} & Text classification & 204 \\
\bottomrule
\end{tabular}
}
\caption{Factors considered in related works and this work. %Factors distinct to our work are shown in bold.
}
\label{table:related_works}
\end{table*}

Our work takes a deeper look at the various factors under several experimental settings. Our key contributions are as follows:
\begin{itemize}[leftmargin=*]
    \item We conduct a comprehensive evaluation of 6  MLLMs, including MLM, autoregressive, and instruction-tuned LLMs, on a text classification task spanning a wide range of languages. This diverse set of models includes \texttt{mBERT} \cite{devlin-etal-2019-bert}, \texttt{XLM-R} \cite{conneau-etal-2020-unsupervised}, \texttt{GPT-3.5} \cite{brown2020language}, \texttt{Bloom} \cite{scao2022bloom} in 5 sizes, \texttt{Bloomz} \cite{muennighoff2023crosslingual} in 5 sizes, and \texttt{XGLM} \cite{lin2022fewshot} in 4 sizes. Additionally, we consider three training scenarios: zero-shot, 2-shot, and fully supervised. 
    \item We consider four key factors in our analysis: pretraining data size, general resource availability levels, language family, and script type. This allows for a more nuanced understanding of the factors influencing MLLM performance.
    \item We leverage the recently introduced SIB-200 dataset \cite{adelani2023sib}, which includes 204 languages, enabling us to investigate MLLM performance across a diverse and extensive linguistic landscape. Between the languages pertaining to the  models and the dataset, we are able to further distinguish them along the dimensions of SEEN, UNSEEN, or ALL, depending on whether the languages were seen during pretraining, or unseen during pretraining, or the set of all languages available in the evaluation dataset, respectively. 
\end{itemize}

By analyzing these factors across different models and training setups, we aim to provide deeper insights into the development of effective and equitable MLLMs for a truly multilingual NLP landscape.

\section{Related Work}

Multilingual NLP research has flourished in recent years, with the development and evaluation of numerous multilingual language models trained on diverse and extensive language datasets. Notable examples include \texttt{mBERT} \cite{devlin-etal-2019-bert}, \texttt{XLM-R} \cite{conneau-etal-2020-unsupervised}, \texttt{mBART} \cite{10.1162/tacl_a_00343}, \texttt{mT5} \cite{xue-etal-2021-mt5}, \texttt{BLOOM} \cite{scao2022bloom}, \texttt{GPT-3} \cite{brown2020language}, \texttt{GPT-4} \cite{openai2023gpt4}, \texttt{LLaMA} \cite{touvron2023llama}, \texttt{PaLM} \cite{chowdhery2022palm}, and \texttt{PaLM 2} \cite{anil2023palm}.

Researchers are increasingly interested in investigating the factors influencing MLLM performance. \citet{wu2020all} examined the impact of pretraining data size, task-specific data size, and vocabulary size on named entity recognition performance.  \citet{scao2022bloom} explored the correlation between probing performance and factors like language family, task-specific dataset size, and pretraining dataset size for the BLOOM model.  \citet{shliazhko2022mgpt} assessed the impact of language script, pretraining corpus size, and model size on language modeling performance, while  \citet{ahuja2023mega} investigated the influence of tokenizer fertility and pretraining data on MLLM performance.

While these studies provide valuable insights, they often focus on a limited set of languages, primarily due to the historical scarcity of annotated multilingual datasets. {Additionally, research by \citet{blasi-etal-2022-systematic} highlights the significant inequalities in the development and performance of language technologies across the world's languages, with a strong bias towards resource-rich languages like English and other Western European languages. Further exacerbating this issue is the lack of representation for dialects, varieties, and closely-related languages within existing datasets. As noted by \citet{faisal2024dialectbench}, this absence hinders the development of NLP systems capable of effectively handling the nuances of linguistic diversity.
} However, the recent emergence of comprehensive multilingual datasets like SIB-200 \cite{adelani2023sib}, and GLOT500 \cite{imanigooghari-etal-2023-glot500} offers exciting opportunities for more extensive and nuanced analyses. Table \ref{table:related_works} summarizes the factors considered in related works and our study. {For a more comprehensive overview of contributing factors to cross-lingual transfer in multilingual language models, readers are encouraged to refer to the review by \citet{philippy-etal-2023-towards}.}

\begin{comment}
    
Our work builds upon these prior efforts by conducting a more holistic investigation with several key distinctions:
\begin{itemize}[leftmargin=*]
    \item Wider Model Coverage: We analyze 17 MLLMs, encompassing both MLM and autoregressive architectures, as well as instruction-tuned LLMs.
    \item More Extensive Language Coverage: We leverage the SIB-200 dataset, covering 204 languages.
    \item Examining Additional Factors: We consider four key factors: pretraining data size, general resource availability levels, language family, and script type.
\end{itemize}
By combining these elements, our work aims to provide deeper insights into the factors that contribute to effective and equitable MLLMs, paving the way for further advancements in multilingual NLP.
\end{comment}

\section{Methodology}

Several factors can influence the performance of multilingual models. In this section, we briefly describe the distinct factors related to typology and data, the dataset of more than 200 languages used for evaluation, and the models we consider in this study.

\subsection{Typology and Data Factors}
We consider various factors to understand their impact on model performance including:

\begin{itemize}[leftmargin=*]

 \item \textbf{Pretraining Data Size:} This refers to the percentage of language-specific data used during the pretraining of each model\footnote{We obtained the train dataset distribution values for \texttt{mBERT} from \url{https://github.com/mayhewsw/multilingual-data-stats} and for \texttt{GPT-3.5} we use proxy statistics from \url{https://github.com/openai/gpt-3/blob/master/dataset_statistics/languages_by_word_count.csv}. Distribution of train dataset for \texttt{XLM-R}, \texttt{BLOOM}, \texttt{BLOOMZ} and \texttt{XGLM} were obtained from their respective papers.}.

  \item \textbf{General Resource Availability (Res Level):} Beyond model-specific resources such as pretraining data size, we also consider a more general notion of resource availability, as per the linguistic diversity taxonomy which categorizes languages into six resource levels \cite{joshi-etal-2020-state}, where level 0 corresponds to low-resource and level 5 corresponds to high-resource level languages. This classification helps us understand the influence of more general resource availability on model performance, and may serve as a proxy when model-specific statistics may not be available (such as in the case of proprietary models). {
     Language resource levels generally correlate positively with models pretraining data sizes, with varying degrees of alignment across different models: mBERT (0.52) and XLM-R (0.48) exhibit relatively stronger correlations, while GPT-3 (0.18), BLOOM (0.37), and XGLM (0.31) show comparatively weaker associations.}

  \item \textbf{Language Family (Lang Family):} The language families that the languages belong to capture some of their linguistic relationships. The information was sourced from the Ethnologue\footnote{\url{https://www.ethnologue.com}} \cite{Ethnologue_2022}.

  \item \textbf{Script:} The script of a language refers to the writing system it employs. This information was sourced from ScriptSource\footnote{\url{https://www.scriptsource.org}}.

\end{itemize}

\subsection{Data}
We systematically study the multilingual models under an important NLP task -- text classification \cite{chang2023language}. The SIB-200 dataset \cite{adelani2023sib} offers a valuable resource for evaluating MLLM performance in a large-scale text classification task,  enabling simultaneous analysis of approximately 200 languages, with text samples categorized into one of seven classes. F1 score is used as the metric for this task.

\begin{figure}[!t]
\centering
\includegraphics[width=0.5\textwidth]{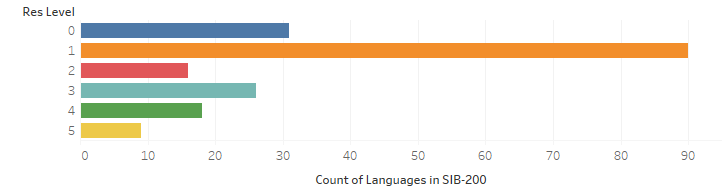}
\caption{Distribution of resource levels in SIB-200.}
\label{figure:reslevel}
\end{figure}

Exploratory analysis of the dataset reveals several interesting insights:
\begin{itemize}
    \item As shown in Figure~\ref{figure:reslevel}, most languages in SIB-200 are classified as resource level 1, indicating a deliberate focus on low-resource languages. This allows us to assess how MLLMs perform on languages with limited linguistic resources available.
    \item Figure~\ref{figure:lang} in Appendix~\ref{app:A} illustrates the distribution of language families within the SIB-200 dataset. Notably, the dataset encompasses 23 different language families, providing a rich linguistic landscape for our analysis. Indo-European languages constitute a significant portion (approximately 36\%) of SIB-200, reflecting their status as the most widely spoken language family globally \cite{Ethnologue_2022}. However, Niger-Congo, Afro-Asiatic, and Austronesian languages also have considerable representation in the dataset. This diverse language family distribution enables us to analyze MLLM performance across different linguistic groups. 
    \item The SIB-200 dataset encompasses text samples written in 29 different script types, offering a diverse range of writing systems for our analysis. As shown in Figure~\ref{figure:script} in Appendix~\ref{app:A}, the Latin script, used by nearly 70\% of the global population \cite{Vaughan_2020}, is the most prevalent writing system in the dataset, followed by Arabic and Cyrillic scripts. This distribution allows us to investigate the impact of script type on MLLM performance.
\end{itemize}

For all evaluations, we use the default train and test splits recommended by the SIB-200 authors. This ensures consistency and comparability across different models and training settings.
%By analyzing MLLM performance on the SIB-200 dataset under these diverse conditions, we aim to gain valuable insights into the factors that contribute to their effectiveness across a wide range of languages and resource levels.

\subsection{Models}
We study the following 6 multilingual language models  spanning various architectures and sizes:

    \begin{itemize}[leftmargin=*]
        \item Masked Language Models (MLMs):
        \begin{itemize}
            \item \texttt{mBERT} (bert-base-multilingual-cased) \cite{devlin-etal-2019-bert}
            \item \texttt{XLM-R} (xlm-roberta-base) \cite{conneau-etal-2020-unsupervised}
        \end{itemize}
        \item Autoregressive Language Models
        \begin{itemize}
            \item \texttt{GPT-3.5} (text-davinci-003) \cite{brown2020language}
            \item \texttt{Bloom} \cite{scao2022bloom} in 5 sizes (560m, 1.1b, 1.7b, 3b, and 7.1b parameters)
            \item \texttt{XGLM} \cite{lin2022fewshot} in 4 sizes (564m, 1.7b, 2.9b, and 7.5b parameters)
        \end{itemize}
        \item Instruction-tuned LLMs:
        \begin{itemize}
            \item \texttt{Bloomz} \cite{muennighoff2023crosslingual} in 5 sizes (560m, 1.1b, 1.7b, 3b, and 7.1b parameters)
        \end{itemize}
    \end{itemize}

    \begin{figure*}[!t]
     \centering    \includegraphics[width=0.94\textwidth]{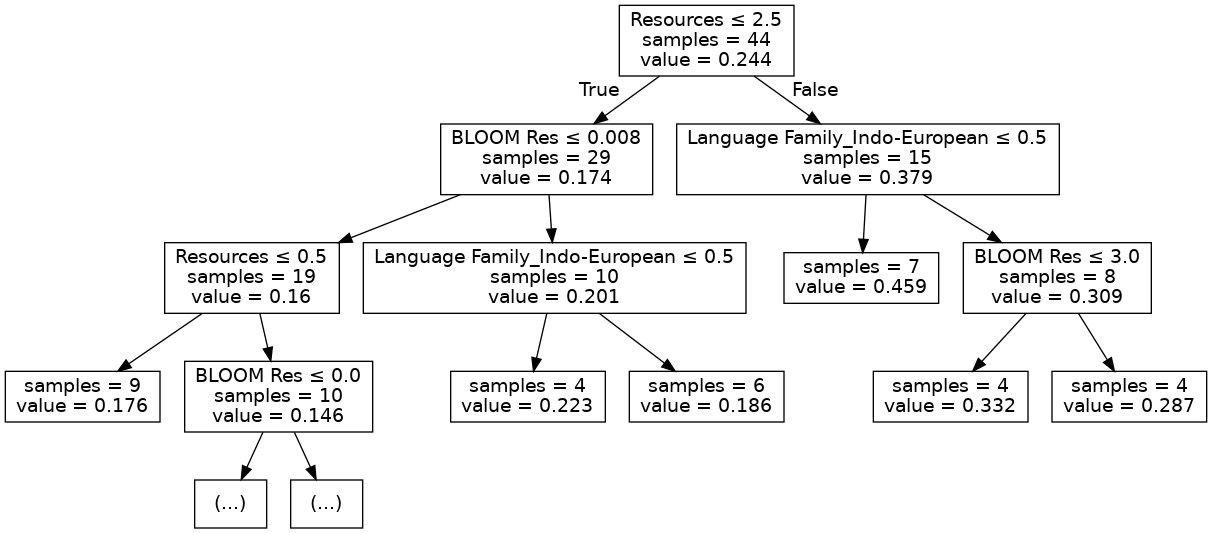}
     \caption{ Decision tree for Bloom-560m (zero-shot, SEEN languages). ``General resource level`` emerges as the most important feature, with a significant performance difference between languages above and below the 2.5 threshold ($p$ < 0.001 as per Mann-Whitney U test).}
     \label{gpt3-SEEN-body}
\end{figure*}

These models were chosen for several key reasons:
\begin{enumerate}
    \item These models provide broad language coverage, allowing us to analyze performance across a diverse set of languages and maximize the linguistic diversity in our study. %This is crucial for understanding how different factors impact MLLM performance across various linguistic contexts.
    \item By including MLMs, autoregressive models, and instruction-tuned LLMs, we can investigate how different model architectures influence performance. %This allows us to assess whether certain architectures are better suited for specific languages or tasks.
    \item The inclusion of models with varying parameter sizes allows us to investigate the interplay between model scale and the factors influencing performance. %This is crucial for understanding whether and how the impact of these factors changes with increasing model size across different languages and tasks.
    \item     \texttt{mBERT} and \texttt{XLM-R}, despite being relatively smaller models, have demonstrated competitive performance even compared to larger models like \texttt{ChatGPT} after fine-tuning \cite{lai2023chatgpt,zhu2023multilingual}. %This makes them valuable models for investigating the factors that contribute to effective multilingual performance.
    \item  The inclusion of both \texttt{Bloom} and \texttt{XGLM}, both autoregressive models, allows us to investigate the impact of pretraining data composition. \texttt{Bloom} focuses more on low-resource languages during pretraining, whereas \texttt{XGLM} emphasizes high-resource languages. This deliberate selection enables us to analyze how the distribution of languages in the pretraining data affects performance across different resource levels.
\end{enumerate}

Note that we primarily focus on models that are open-source or have made the list of pretraining languages and data composition available. 

Additionally, we consider the following training and inference scenarios:
\begin{itemize}
   
    \item Zero-shot: \texttt{GPT-3.5}, \texttt{Bloom}, \texttt{Bloomz}, and \texttt{XGLM} were evaluated directly on the test set without any specific fine-tuning. This assesses the model's ability to generalize to unseen tasks and languages based on its pretrained knowledge.
    \item Two-shot In-Context Learning (ICL): \texttt{Bloom}, \texttt{Bloomz}, and \texttt{XGLM} were also evaluated in  two-shot ICL setting where the models were provided with two labeled examples for each class from the train set. This allows us to particularly investigate effective factors  for improving performance of unseen languages. {We opted for two demonstrations in ICL to keep the input length shorter than the context length of our models across all languages. } %This decision was influenced by both preliminary experiments and existing research, which show that even a small number of examples can effectively help LLMs adapt to new tasks and languages \cite{brown2020language}.
     \item Full-shot: \texttt{mBERT} and \texttt{XLM-R} were fine-tuned on the SIB-200 training set and evaluated on the test set. %This represents a supervised learning setting where models are explicitly trained on the target task and languages.
\end{itemize}

For full-shot training of mBERT and XLM-R, we adhered to the hyperparameters recommended by the SIB-200 paper authors to ensure consistency with the original dataset benchmarks. For Bloom, Bloomz, and XGLM in both zero-shot and two-shot ICL settings, as well as for GPT-3.5 in zero-shot setting, we use prompts to frame the text classification task, which are detailed in Appendix~\ref{app:prompts}.

%By analyzing this diverse set of MLLMs, we aim to gain a comprehensive understanding of the factors influencing their performance across a wide range of languages and contribute to the development of more effective and equitable multilingual NLP systems.

\section{Results and Analysis}
Now we discuss the results of our comprehensive experiments. We focus on analyzing the performance of models across three distinct scenarios: ALL, SEEN, and UNSEEN. The ALL scenario considers all languages in the SIB-200 dataset for which resource level information is available\footnote{This information is available for 190 languages.}. The SEEN scenario focuses on languages included in the pretraining data of the respective MLLMs, while the UNSEEN scenario examines performance on languages not present in the pretraining data. %Within each scenario, we analyze the impact of various factors on MLLM performance, including model-specific resources, general resource availability levels, language family, and script type. %By examining these factors and their interactions across different scenarios, we aim to provide a comprehensive and nuanced understanding of the factors that contribute to effective MLLM performance across diverse languages and resource levels.

In total, results are obtained from 93 distinct experimental settings (models of different sizes, training scenarios, and language categories of seen/unseen/all).

To understand the complex interplay of multiple factors influencing MLLM performance, we employ decision tree analysis for statistical inference. This approach is well-suited for handling factors of different types, including categorical, ordinal, and numeric data. Decision trees are trained to predict the F1 score of models based on language features. By analyzing the resulting tree structure, we can gain insights into the relative importance of different features and their interactions.

{As decision trees were trained on the entirety of our data, traditional methods for testing their performance were not applicable. Instead, we employed the Mann-Whitney U test \cite{mann1947test}, to ensure that the features appearing at the root of the decision trees were indeed relevant and contributed significantly to the differentiation between the language splits. This approach allowed us to validate the significance of the features identified by the decision tree in delineating distinct language groups without relying solely on the performance metrics of the decision tree models themselves.}

% Following the decision tree analysis, we utilize the Mann-Whitney U test \cite{mann1947test} to determine statistically significant differences in performance between groups identified by the decision tree.

\begingroup\tabcolsep=1pt\def\arraystretch{1}
\small
\begin{table*}[!t]
    \centering
    \begin{tabular}{l|c|c|c}
    \cline{2-4}
         \multicolumn{1}{c|}{}&\multicolumn{3}{|c|}{\bf Zero-shot}\\   
         \midrule
         Model&  ALL&  SEEN&  UNSEEN\\  
         \midrule
          \texttt{Bloom-560m}&   Pretrain data (<=0.125\%)&   Resource level (<=2.5)&   Script (Latin or not)\\   
         \texttt{Bloom-1b1}&  Pretrain data (<=0.125\%)&  Resource level (<=2.5)&  Script (Devanagari or not)\\   
         \texttt{Bloom-1b7}&  Pretrain data (<=0.175\%)&  Resource level (<=2.5)&  Script (Latin or not)\\   
         \texttt{Bloom-3b}&  Pretrain data (<=0.175\%)&  Resource level (<=2.5)&  Script (Latin or not)\\   
         \texttt{Bloom-7b1}&  Pretrain data (<=0.125\%)&  Resource level (<=2.5)&  Script (Devanagari or not)\\   
         \texttt{Bloomz-560m}&  Script (Latin or not)&  Pretrain data (<=0.03\%)&  Script (Latin or not)\\   
         \texttt{Bloomz-1b1}&  Pretrain data (<=0.008\%)&  Pretrain data (<=0.03\%)&  Script (Latin or not)\\   
         \texttt{Bloomz-1b7}&  Pretrain data (<=0.008\%)&  Pretrain data (<=0.03\%)&  Script (Latin or not)\\ 
 \texttt{Bloomz-3b}& Pretrain data (<=0.002\%)& Pretrain data (<=0.013\%)& Script (Latin or not)\\
 \texttt{Bloomz-7b1}& Pretrain data (<=0\%)& Pretrain data (<=0.9\%)& Script (Latin or not)\\
 \texttt{XGLM-564m}& Pretrain data (<=0.003\%)& Resource level (<=2)& Lang. family (Austronesian or not)\\
 \texttt{XGLM-1.7b}& Pretrain data (<=0.006\%)& Pretrain data (<=1.487\%)& Script (Devanagari or not)\\
 \texttt{XGLM-2.9b}& Pretrain data (<=0.003\%)& Script (Latin or not) & Script (Devanagari or not)\\
 \texttt{XGLM-7.5b}& Pretrain data (<=0\%)& Pretrain data (<=1.122\%)& Script (Devanagari or not)\\
 \texttt{GPT-3.5}& Resource level (<= 2.5)& Pretrain data (<=5.312\%)& Lang. family (Indo-European or not)\\
 \addlinespace
 \cline{2-4}
 \multicolumn{1}{c|}{} & \multicolumn{3}{|c|}{\bf Two-shot ICL}\\
 \midrule
 Model & ALL& SEEN& UNSEEN\\
 \midrule
 \texttt{Bloom-560m}& Pretrain data (<=0.045\%)& Pretrain data (<=0.045\%)& Lang. family (Indo-European or not)\\
 \texttt{Bloom-1b1}& Pretrain data (<=0.095\%)& Pretrain data (<=0.095\%)& Script (Latin or not)\\
 \texttt{Bloom-1b7}& Pretrain data (<=0.175\%)& Pretrain data (<=0.175\%)& Script (Latin or not)\\
 \texttt{Bloom-3b}& Pretrain data (<=0.008\%)& Pretrain data (<=0.008\%)& Script (Latin or not)\\
 \texttt{Bloom-7b1}& Pretrain data (<=0.008\%)& Pretrain data (<=0.008\%)& Script (Latin or not)\\
 \texttt{Bloomz-560m}& Pretrain data (<=0.03\%)& Pretrain data (<0.03\%)& Script (Devanagari or not)\\
 \texttt{Bloomz-1b1}& Pretrain data (<=0.008\%)& Pretrain data (<=0.013\%)& Script (Latin or not)\\
 \texttt{Bloomz-1b7}& Pretrain data (<=0.005\%)& Pretrain data (<=0.013\%)& Script (Cyrillic or not)\\
 \texttt{Bloomz-3b}& Pretrain data (<=0\%)& Pretrain data (<=0.9\%)& Script (Latin or not)\\
 \texttt{Bloomz-7b1}& Pretrain data (<=0\%)& Pretrain data (<=0.013\%)& Script (Latin or not)\\
 \texttt{XGLM-564m}& Pretrain data (<=0.003\%)& Pretrain data (<=0.095\%)& Lang. family (Niger-Congo or not)\\
 \texttt{XGLM-1.7b}& Pretrain data (<=0.003\%)& Resource level (<=2)& Script (Devanagari or not)\\
 \texttt{XGLM-2.9b}& Pretrain data (<=0.003\%)& Script (Latin or not)& Lang. family (Indo-European or not)\\
 \texttt{XGLM-7.5b}& Pretrain data (<=0.003\%)& Pretrain data (<=0.15\%)& Lang. family (Indo-European or not)\\
 \addlinespace
 \cline{2-4}
 \multicolumn{1}{c|}{} & \multicolumn{3}{|c|}{\bf Full-shot}\\
 \midrule
 Model & ALL& SEEN&UNSEEN\\
 \midrule
 \texttt{mBERT} &  Pretrain data (<=3.786\%)&  Pretrain data (<=8.627\%)& Lang. family (Indo-European or not)\\
 \texttt{XLM-R}& Pretrain data (<=13.5\%)& Pretrain data (<=90\%)&Lang. family (Indo-European or not)\\ 
 \bottomrule
    \end{tabular}
    \caption{Top features identified by decision tree analysis for each model and scenario. For SEEN languages, pretraining data size and resource level dominate (except for \texttt{XGLM-2.9b}, where script type is most influential). For UNSEEN languages, linguistic characteristics (script type and language family) take precedence. All features exhibit statistically significant differences in performance ($p$ < 0.001).}
    \label{tab:top_features}
\end{table*}
\endgroup

Figure~\ref{gpt3-SEEN-body} presents the decision tree analysis for the \texttt{Bloom-560m} model on SEEN languages, revealing {\em general resource level} as the most influential feature. Specifically, the tree distinguishes between languages with resource levels below 2.5 (levels 0,1,2) and those above 2.5 (levels 3,4,5). Among the 44 SEEN languages, the 29 languages with resource levels below 2.5 exhibit a mean F1 score of 0.174, while the 15 languages with higher resource levels achieve a significantly higher mean F1 score of 0.379. A Mann-Whitney U test confirms a statistically significant difference in performance between these two groups (p < 0.001). This suggests that for the \texttt{Bloom-560m} model on SEEN languages, the general resource level of a language plays a crucial role in determining its performance, with higher resource levels leading to better performance.
By employing this combined approach of decision tree analysis and statistical testing, we can effectively disentangle the complex relationships between various factors and their impact on MLLM performance.

\begin{figure*}
          \centering
     \includegraphics[width=\textwidth]{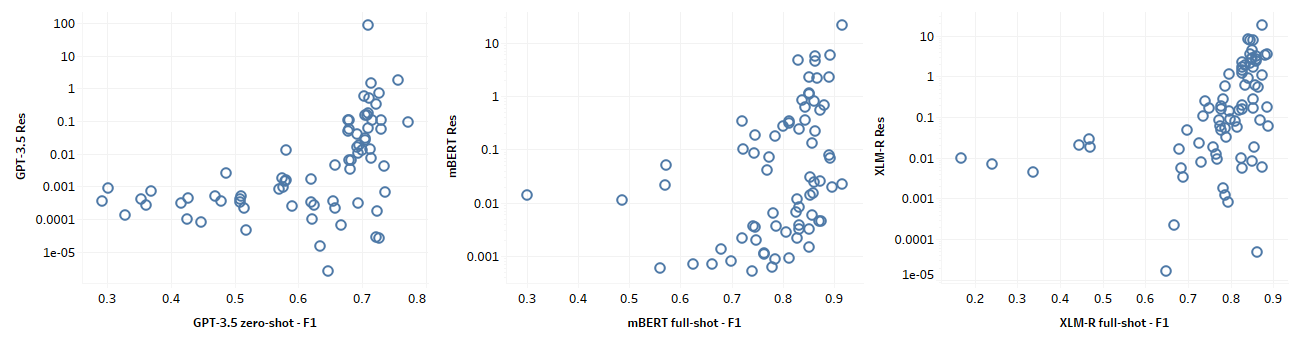}
     \caption{F1 Score vs. model-specific pretraining data (percentage) for GPT-3.5, mBERT and XLM-R models.}
     \label{3model-res}
\end{figure*}

The summarized results\footnote{Detailed decision trees for all models and setups are available in our repository: \url{https://github.com/PortNLP/MLLMs_performance}} of all 93 decision tree analyses are presented in Table~\ref{tab:top_features}. We observe distinct patterns in feature importance across the three scenarios:

\medskip
\medskip
\noindent \textbf{\textit{ALL Languages:}}
\begin{itemize}[leftmargin=*]
    \item For the ALL languages scenario, decision trees clearly reveal that pretraining data is the most influential factor in 29 out of 31 cases. Because ALL includes languages SEEN and UNSEEN, notably, our deeper look at the decision tree analyses indicates that this factor in most cases boils down to {\em whether the language was part of the training set or not, rather than the amount of language-specific data}, as indicated by the values of the pretraining data percentages which range from 0\% to at most 0.175\%. GPT-3.5 model draws the distinction along general resource levels whether a language is low resource (0, 1, or 2) or level 3 and higher. 
\end{itemize}

\medskip
\noindent \textbf{\textit{SEEN Languages:}}
\begin{itemize}[leftmargin=*]
    \item For SEEN languages, model-specific pretraining data continues to remain the most influential factor in 22 out of 31 model and scenario combinations. However, this time because there are no unseen languages in the mix, the model performance seems to be impacted by the amount of  pretraining data, as indicated by the slightly higher percentage values as compared to the ALL languages scenario.
    \item Interestingly, general resource availability based on linguistic diversity taxonomy \cite{joshi-etal-2020-state} appears to be the most important factor for \texttt{Bloom} models in the zero-shot setup, as well as for \texttt{xglm-564m} (zero-shot) and \texttt{xglm-1.7b} (two-shot). For \texttt{Bloom} models, the distinction is along resource levels 0/1/2 or higher, whereas for \texttt{xglm} models, it is along 0/1 and higher. Additionally, \texttt{xglm-2.9b} in both zero-shot and two-shot scenarios shows a stronger influence of script type (Latin or not). These cases indicate that factors beyond pretraining data size can also play a significant role for specific models and settings.
    \item Furthermore, Figure \ref{3model-res} plots the performance of \texttt{mBERT}, \texttt{XLM-R}, and \texttt{GPT-3.5} models in relation to model-specific pretraining data amounts. The figure demonstrates a clear trend: as the model-specific language data increases, so does the model's performance. This observation aligns with the finding that pretraining data size is a crucial factor for SEEN languages. %Similar patterns are observed for other models and scenarios, included in Appendix \ref{app:A}.
\end{itemize}

\medskip
\noindent \textbf{\textit{UNSEEN Languages:}}
\begin{itemize}[leftmargin=*]
    \item In contrast to SEEN languages, UNSEEN languages show quite a different pattern. Naturally, because UNSEEN languages do not have pretraining data as one of their relevant factors, it is absent from this column. However, out of 31 models, 23 are most impacted by script type, and 8 are most influenced by language family. This shift in importance towards linguistic features suggests that when models encounter unfamiliar languages, they rely more heavily on similarities in writing systems to generalize from their existing knowledge. 
    \item Within the scripts and language families, there are nuanced differences. For instance, while generally the models make the distinction along the lines of whether the script is Latin or not, occassionally Devanagari script also seems important, particularly for \texttt{XGLM} models. Similarly, while Indo-European is the most common influential language family, we also observe an instance each of Austronesian and Niger-Congo. Additionally, models of different sizes from the same family may prefer not just a different script or a different language family when moving from zero-shot to two-shot setting, they may prefer an entirely different factor (e.g., \texttt{Bloom-560m} in zero-shot vs. two-shot settings), further complicating the matters. 
    %\item The lack of influence from general resource availability and the emphasis on script and language family suggest that MLLM performance on UNSEEN languages is primarily driven by cross-lingual transfer learning. Models seem to perform better on UNSEEN languages that share script or language family characteristics with languages they have seen during training.
\end{itemize}

\section{Discussion}

Our comprehensive analysis of 6 multilingual models on the SIB-200 dataset reveals valuable insights into the factors influencing their performance across a diverse range of languages.

%These findings highlight the complex interplay of factors influencing MLLM performance and how this interplay varies depending on the language's familiarity to the model. While pretraining data size plays a dominant role for SEEN languages, script type and language family become more crucial for UNSEEN languages, suggesting the importance of cross-lingual transfer learning. For the overall performance across all languages, the presence or absence of a language in the training data emerges as the most significant factor.

%This analysis provides valuable insights into the strengths and limitations of current MLLMs and can guide future research in developing models that perform more effectively and equitably across diverse languages, including those with limited resources and representation in training data.

{Our key findings can be summarized as follows:}

\begin{itemize}
    \item Pretraining data size consistently emerges as a crucial factor, but the distinction is less along the quantity of data but rather whether the languages have been  encountered during training or not. %and this effect becomes more pronounced with larger pretraining datasets.
    \item For UNSEEN languages, script type and language family are influential, suggesting that MLLMs rely on cross-lingual transfer learning to generalize to unfamiliar languages. %Models perform better on UNSEEN languages that share script or language family characteristics with languages they have seen during training.
    \item General resource availability plays a less prominent role overall but appears to be important for one specific model under one setting (\texttt{Bloom} in zero-shot for seen languages). %specific models and settings, particularly in the zero-shot scenario.
    \item Interestingly, the performance of \texttt{Bloomz}, an instruction-tuned model, is more influenced by the distribution of languages in its pretraining corpus than the fine-tuned dataset used for instruction tuning. This suggests that the initial pretraining stage plays a crucial role in shaping the model's capabilities, even after further fine-tuning for specific tasks.
    \item Finally, our analysis also indicates that while model size and architecture may influence overall performance, they do not significantly alter the most important features identified by the decision trees. The distribution of languages in the pretraining data and the linguistic characteristics of the target languages consistently emerge as the dominant factors regardless of the specific model architecture or scale.
\end{itemize}

%These findings highlight the complex interplay of factors affecting MLLM performance and how this interplay varies depending on the language's familiarity to the model. While increasing pretraining data size is beneficial, it might not be sufficient for achieving optimal performance on unseen languages. \\

Several future directions remain to be explored. We observed that script type can be more influential for specific models and settings. Further investigation is needed to understand the reasons behind these preferences and how they can be leveraged to achieve more consistent performance across languages. It is also not clear why models lean towards different factors under different settings (for instance, resource level is important in \texttt{Bloom-560m} zero-shot setting but pretraining data is important in its two-shot ICL setting).

%further research should explore methods to enhance cross-lingual transfer learning in MLLMs. This might involve incorporating linguistic knowledge, developing new training strategies, or exploring alternative features that capture cross-lingual similarities more effectively. Addressing Resource Imbalance: The disparity in performance between high- and low-resource languages remains a significant challenge. Future research should focus on developing techniques to improve MLLM performance on low-resource languages, potentially by leveraging transfer learning from high-resource languages or exploring data augmentation strategies. 

\section{Conclusion}

This study analyzed 6 multilingual language models on the SIB-200 dataset, revealing key insights into their performance across around 200 languages. We found that the size of the pretraining data significantly affects performance. For unseen languages, script type and language family become more crucial, highlighting the importance of cross-lingual transfer learning. While general resource availability plays a less prominent role overall, it can be significant for specific models and settings. %Notably, instruction-tuned models like \texttt{BLOOMZ} are still heavily influenced by the distribution of languages in their pretraining data, even after further fine-tuning. 
Interestingly, model size and architecture do not significantly change the most important features identified in our analysis. Our work contributes to a deeper understanding of MLLMs and hopes to guide the development of more effective and equitable multilingual NLP systems.

%These findings emphasize the complex interplay of factors affecting MLLM performance and underscore the enduring influence of pretraining data. Future research should prioritize improving cross-lingual transfer learning and incorporating linguistic knowledge to enhance performance across diverse languages. 

\section*{Limitations}
This study provides insights into multilingual language model performance, but it is important to acknowledge certain limitations. The SIB-200 dataset, while extensive, may contain biases in language representation and genre distribution, potentially affecting the generalizability of our findings. Additionally, our analysis focuses on the text classification task, and the findings may not directly generalize to other NLP tasks. While we analyzed a diverse set of models, our findings may not be fully representative of the entire MLLM landscape. Finally, our analysis is based on the current state of MLLMs, and the relative importance of different factors may change as these models continue to evolve. Future research should address these limitations by expanding to more diverse datasets, investigating different NLP tasks, evaluating a broader range of models, and conducting longitudinal studies.

\section*{Ethics Statement}

The experimental setup and code implementation ensured adherence to ethical guidelines, data usage agreements, and compliance with the terms of service of the respective language models and data sources. The research team also recognized the importance of inclusivity and fairness by considering a diverse set of languages and language families in the evaluation, thus avoiding biases and promoting balanced representation.

\section*{Acknowledgements}
{We are grateful to the anonymous reviewers whose feedback and thought-provoking questions  enhanced this paper. The engaging discussions and collaborative spirit within the PortNLP research group were instrumental in shaping this research. We acknowledge the National Science Foundation for their financial support through grants (CRII:RI 2246174 and SAI-P 2228783), which made this work possible.}
%\section*{Acknowledgements}

\FloatBarrier
\bibliography{anthology,custom}

\appendix

\section{Appendix: Prompts} \label{app:prompts}
This appendix provides the specific prompts used for evaluating \texttt{Bloom}, \texttt{Bloomz}, \texttt{XGLM}, and \texttt{GPT-3.5} in the zero-shot and two-shot in-context learning (ICL) settings on the SIB-200 text classification task.\\

\textbf{Zero-shot Prompt (Bloom, Bloomz, XGLM):}
\begin{verbatim}
SENTENCE: “{input sentence}”
Is this SENTENCE science, travel, politics,
sports, health, entertainment, geography?
OPTIONS:
-science
-travel
-politics
-sports
-health
-entertainment
-geography
ANSWER:
\end{verbatim}

\textbf{Two-shot ICL Prompt (Bloom, Bloomz, XGLM):}
\begin{verbatim}
What category does SENTENCE belong to?

SENTENCE: “{sentence1}”
LABEL: {label1}
SENTENCE: “{sentence2}”
LABEL: {label2}
...
SENTENCE: “{sentence14}”
LABEL: {label14}
SENTENCE: “{input sentence}”
OPTIONS:
-science
-travel
-politics
-sports
-health
-entertainment
-geography

LABEL:
\end{verbatim}

\textbf{Zero-shot Prompt (GPT-3.5):}
\begin{verbatim}
You will be provided with a text, and your task
is to classify its category as science, travel,
politics, sports, health, entertainment,
geography.
{input sentence}

Category: 

\end{verbatim}

\section{Appendix: Supplemental plots} \label{app:A}

\begin{figure*}[!htbp]
\centering
\includegraphics[width=0.7\textwidth]{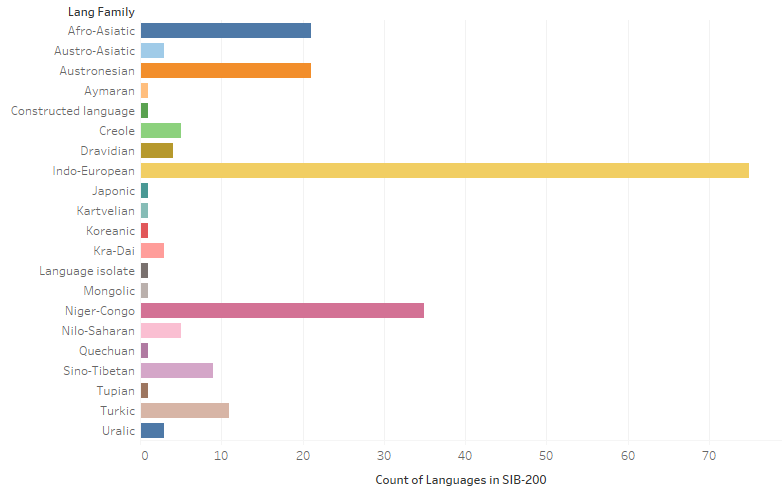}
\caption{Distribution of language family in SIB-200.}
\label{figure:lang}
\end{figure*}

\begin{figure*}[!htbp]
\centering
\includegraphics[width=0.6\textwidth]{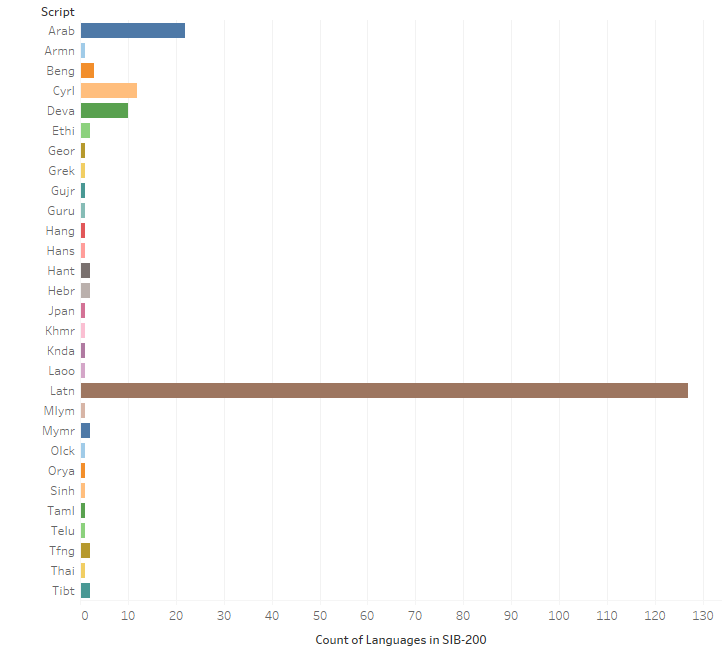}
\caption{Distribution of scripts in SIB-200.}
\label{figure:script}
\end{figure*}

\end{document}